# Enhanced Facial Recognition Framework based on Skin Tone and False Alarm Rejection

Ali Sharifara[1], Mohd Shafry Mohd Rahim[2], Farhad Navabifar[2], Dylan Ebert[1], Amir Ghaderi[1], Michalis Papakostas[1]

[1] University of Texas at Arlington (UTA), Arlington, USA
[2] University of Technology (UTM), Skudai, Malaysia
ali.sharifara@uta.edu, Shafry@utm.my, nfarhad49@gmail.com, dylan.ebert@mavs.uta.edu,
amir.ghaderi@mavs.uta.edu, michalis.papakostas@mavs.uta.edu

## ABSTRACT
Face detection is one of the challenging tasks in computer vision. Human face detection plays an essential role in the first stage of face processing applications such as face recognition, face tracking, image database management, etc. In these applications, face objects often come from an inconsequential part of images that contain variations, namely different illumination, poses, and occlusion. These variations can decrease face detection rate noticeably. Most existing face detection approaches are not accurate, as they have not been able to resolve unstructured images due to large appearance variations and can only detect human faces under one particular variation. Existing frameworks of face detection need enhancements to detect human faces under the stated variations to improve detection rate and reduce detection time. In this study, an enhanced face detection framework is proposed to improve detection rate based on skin color and provide a validation process. A preliminary segmentation of the input images based on skin color can significantly reduce search space and accelerate the process of human face detection. The primary detection is based on Haar-like features and the Adaboost algorithm. A validation process is introduced to reject non-face objects, which might occur during the face detection process. The validation process is based on two-stage Extended Local Binary Patterns. The experimental results on the CMU-MIT and Caltech 10000 datasets over a wide range of facial variations in different colors, positions, scales, and lighting conditions indicated a successful face detection rate.

## 1. INTRODUCTION
In recent decades, human face detection has been researched widely. The main reason comes from the recent advances of its applications such as video surveillance system, security access control, information retrieval in many unstructured multimedia databases, and advanced Human Computer Interaction (HCI) [1]. The input images can be captured via several devices such as cameras and they can be manipulated by various computer vision methods [1, 2]. In addition, most of the biometric and HCI applications include computing analysis on human faces such as in face alignment, recognition, verification, and authentication purposes. Indeed, human faces must be detected before any such analysis can occur in these images [1]. Integrating information from various visual cues, such as texture, stereo disparity, and image motion, have better improvement in performance on perceptual tasks, such as face detection. On the other hand, the additional determination required to extract and signify information from additional cues may increase computational complexity [17]. Several approaches are proposed to solve this problem, including template matching, knowledge based, Adaboost learning based, Neural Networks, and SVM algorithms. However, success is accomplished with each face detection method with varying degrees and complexities [12]. Most of the face detection frameworks try to extract a portion of the whole face, thus eliminating most of the background [11]. Real-time face detection involves detection of a face from a series of frames from a video capturing device. Although the hardware necessities for such a system are far more inflexible, from a computer vision standpoint, real-time face detection is actually a far simpler process than detecting a face in a static image. This is due to the face, unlike most of our surrounding environment, is frequently moving. Therefore, face detection has an outstanding importance and plays a critical role in most of the face processing systems and the performance of this step has direct impact on the overall performance of the systems [25].

Our paper is organized as follows: First, the current section presents an introduction of the face detection, and the existing challenges involved in this process. Next, it presents some of the existing methods and related work. Then, it presents the proposed face detection and provides the details of the framework, which consists of main contributions to the study, which reduced search-space, and rejects false positives. Then, it provides the experimental results of the detection process, which includes search space reduction, feature extraction, classification, and the validation process. Moreover, in this section, we also present the comparison between the proposed framework and other well-known methods in skin color segmentation and final face detection results based on the available online datasets. Finally, concludes the major achievements in this research.

## 2. RELATED WORK
Automatic face detection is one of the main and significant processes in most face processing applications. In addition, the existing methods have limitations and they are not accurate, especially under certain variations. For example, most of the face detection algorithms are able to locate the face in Figure 1 (a), due to this image being taken in frontal view and in good lighting. To locate these types of images is not a challenging issue. The challenges start when a face detector has to struggle with the second type of images (Figure 1 (b)), which creates all the mentioned limitations that is called unstructured settings. The unstructured settings in images may contain several variations such as complex backgrounds, different lighting conditions, poses, races, skin colors, age, gender, face occlusion etc. [5, 11]. Figure 1 (a) shows an image that contains a face that is easy to be detected, meanwhile Figure 1 (b) depicts an image, which is not easy to detect by a face detector due to several variations such as low contrast, profile-view, and complex backgrounds. These types of images have several limitations and challenging issues.

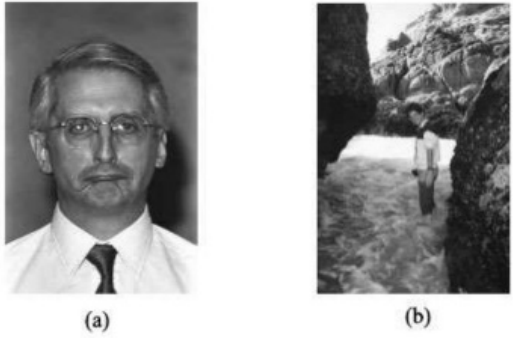

**Figure 1. Sample image from Caltech 10000. (a) Image without challenge (b) Image with challenges.**

The main problem and challenging issue, which is addressed in most of the face detection methods, is "high false alarms", in which some objects are selected incorrectly as human faces and some other objects, which are human faces, are rejected [4]. By increasing the face detection rate, the false positive rate also increased accordingly [23]. Research shows that this phenomenon happens in most of the face detection methods, and this problem becomes worse when some other challenges occur at the same time, such as illumination or occlusion in an image [16]. The problem is illustrated in Figure 2, which shows the correct and incorrect selected faces in an image.

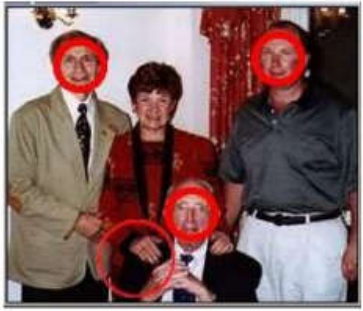

**Figure 2. An example of face detector output with low accuracy.**

To evaluate the performance of face detection, there are several metrics including detection rate, false alarm rate, missing rate, etc. [11]. Most practical applications are able to increase the detection rate, and false alarm rate accordingly. This is due to the decision threshold that is taught to distinguish between face and non-face under training data set [20].

The present research addresses several contributions in digital image processing including image enhancement, skin color segmentation, feature extraction and classification processes. In general, search space reduction is a task to locate and eliminate the regions in the image, which are not contain of human skin tones. This task can be performed by segmentation process. The segmentation process must be able to eliminate the background of input images and select the regions, which contain skin color tones [6]. An appropriate segmentation method can reduce the computation time of the face detection process. In this study, a preprocessing module is employed to remove noise of input image and shrink the search space in order to decrease the computation time for feature extraction and classification procedures. For validation purposes, another post-processing process is applied to confirm or reject the potential faces from previous stage. The validation process is based on the extended local binary pattern and support vector machines. As a result, the main contribution of this study is to keep the detection rate high as well as decreasing the number of false alarms by applying a 3-stage face detection system, which is done based on Adaboost algorithm and modified Haar-like features in the main stage and final validation process.

## 3. PROPOSED METHOD

Face detection is one of the main processes in face processing applications. In addition, the existing methods have some limitations and they are not accurate, especially under certain variations, which have been discussed in the previous section, including illumination, poses, and occlusion. Therefore, to overcome the limitations and increasing the accuracy of face detection problem, there are some essential steps of image processing algorithm, which need to apply on input images. These steps include image enhancement, image segmentation, feature extraction, classification, and validation.

The proposed framework consists of four main phases including preprocessing, search space reduction, detection and the validation process. In each stage, several algorithms are applied and the result of each stage is the input for the next stage. In order to enhance the quality of the image and reduce noise, a preprocessing unit is applied, which consists of several noise reduction methods. Firstly, in the following section, several preprocessing processes are described, in order to remove noise from the input images and make them ready for the next stages of face detection. Then, the details of the proposed methods to optimize the search space and eliminate false alarms are provided in the following sections. In addition, two different stages are developed based on different approaches in order to increase the accuracy and performance of the face detection framework. The proposed framework design is depicted in Figure 3.

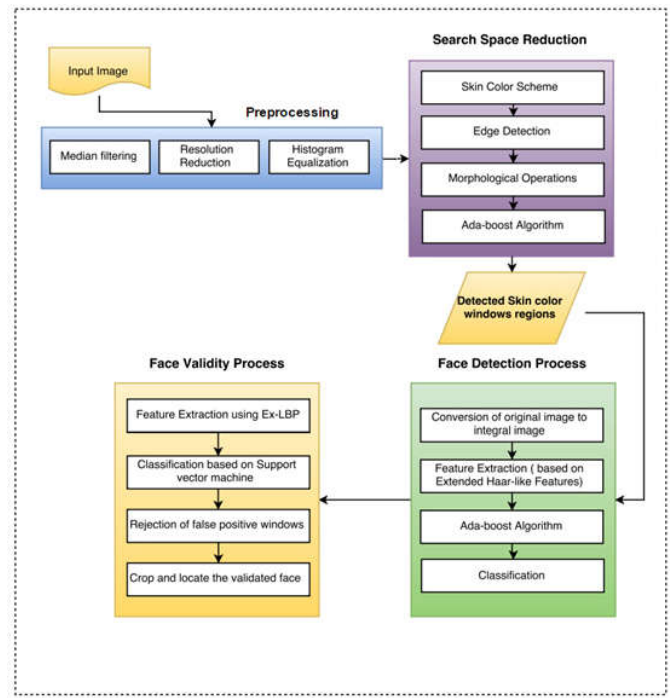

**Figure 3. The proposed framework design.**

The search space reduction consists of several methods, including a skin-color method, edge detection, morphological operations, and the Adaboost algorithm. In addition, the validation process contains four main stages, including feature extraction, classification, rejection of false positive windows, and locating the face in the original image.

### 3.1 Preprocessing Enhancement

The image preprocessing process is an extremely significant stage in the face detection system [19]. Indeed, there are several challenging issues and difficulties, which can add some complexity to the classification process. Hence, they might have an effect on the result of face detection such as noise, brightness, lighting conditions, complex background /texture, pose variation, occlusion, and scaling. In order to avoid this kind of challenge, several preprocessing activities are needed to normalize the input image.

## 3.2 Search Space Reduction

S Search space optimization is the process of selecting region of interest (ROI) from the input image and ignoring the rest of the regions, which is known as the Region of Non-Interest (RONI). This process is performed after the preprocessing phase. The main aim of this step is to ignore the background of an image that does not contain skin color. The process of segmentation initialises with a comparison between the threshold and each pixel. The process subtracts the given threshold from each pixel and verifies whether the result is negative or positive. In case that the result is negative, then the threshold is greater than a certain pixel. The process will be continued with a 2s complement number. Then, an AND gate is deployed to compare the most significant bits of each pixel. In this study, a skin-color segmentation based method is proposed in order to reduce the search space of an input image. The proposed method is based on the human skin color and edges, which only selects the regions that contain the skin color. Final evaluation will be calculated based on Equation 1.

$$\text{Skin Color} = \frac{\sum_{i=1}^{max} SC}{\sum_{i=1}^{max} IP} \times 100 \quad (1)$$

Where, SC is skin-color pixels and IP is total of pixels which exist in the input image. At this point, all of the skin areas, which identified are selected, and proceed to count the amount of pixels that exist within the input image.

### 3.2.1 Skin Color Scheme

Color is one of the most distinguishing features of an image and is applied in many skin and face detection methods [26]. In spite of extensive research efforts in this area, selecting an appropriate color space in terms of skin and face classification has remained an open topic. The performance can address issues like variations in illumination, and variety in skin color tones. Hence, for the purpose of segmentation, a skin color-based segmentation method is proposed which is the combination of Sobel edge detection and the YCbCr color space to detect the skin pixels. The method is proposed in an attempt to reduce the search space of input images and increase computation time of feature extraction and classification in comparison with existing methods. By observing the segmented skin regions, we see that not all of them are related to the human face. A face classification method is used to verify whether the selected regions are part of the human face or not. In the following part, we have described the proposed framework in a few steps:

**Step-1:** Firstly, the chromatic distance values needs to be obtained for the input image. The chromatic distance values help to illustrate the intensity of image, and subsequently, the value of Cr and Cb can be obtained to display in a graph.

**Step-2**: Secondly, in order to get the filtered image, a low pass filter needs to be applied. In addition, RGB input images need to be converted to YCbCr color space.

**Step-3**: In the third step, there is a conversion of input color image to a grayscale image. Then, edge detection using Sobel operators will be done to find the edges of image by an initial threshold. The improvement can be achieved by applying a down scale factor and the image can be shrunk without losing data. Hence, the algorithm executes on less pixels. The shrinking process can be achieved by analyzing each nM column and row, where M is n - 0, 1, 2 ...S and M is the downscale factor. Another enhancement is applied to remove the noise of such holes in the result. Therefore, a closure morphological method can be utilized. Finally, after performing some operations like converting an image into grayscale and grayscale into binary image, the final segmented image can be obtained. Therefore, a skin color based algorithm is proposed in order to reduce the search space and eliminate the windows, which are not contained human skin color. Then, the selected windows are considered as an input of the feature extraction process and classification.

The result of this step is the binary image from the skin-color segmentation of the last stage. Initially, morphological opening operator is used in order to eliminate extremely tiny objects from the image, while keeping the shape and size of large objects in the image [19].

## 3.3 Face Detection Process

The proposed face detection framework consists of several phases. The input of face detection process is the result of image segmentation, which is done before feature extraction process. The feature extraction stage acts as the main component of any pattern recognition system [8]. Raw intensity values, as directly provided by a camera, from the most basic manner of representing an image. It is believed that even a single object can represent a wide range of appearances, as the environment can influence its appearance with shadows, lighting conditions, occlusions, etc. The input of a learning algorithm is features and the most important reason of using features instead of raw pixel values is to decrease the in-class while increasing the out-of-class variability in comparison to the raw data, then make classification easier. In addition, the speed of feature evaluation is a significant factor while most of the face detection algorithms slide a fixed-size window that scales over the input image. Haar-like wavelet features are defined as a variation between unfilled rectangles and the accumulated intensities of filled rectangles. Figure 4 depicts the extended Haar-like features, which significantly enrich the basic set of conventional Haar-like features, and they have proved to obtain better result especially in real time applications [9].

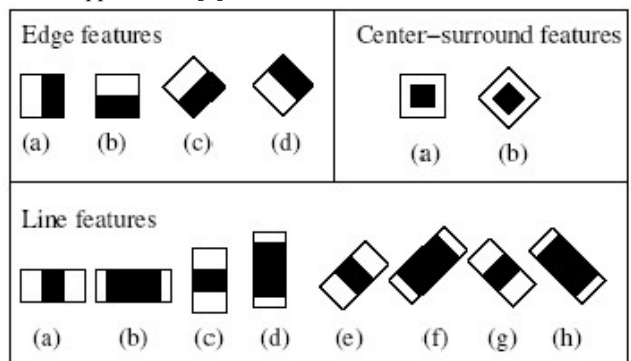

**Figure 4. The extended Haar-like feature set proposed [9].**

The value of local feature *f* is calculated by subtracting white regions from black regions ($w_i$-$b_i$). In order to speed up the process, Integral Image is employed, as discussed earlier. The input of a learning algorithm is features, and the most important reason of using features instead of raw pixel values is to decrease the in-class while increasing the out-of-class variability in comparison with the raw data and then creation classification easier. A number of researchers have proposed different approaches to expand the robustness and discriminative influence of Haar-like features [10]. Isolated value of pixels do not provide any information other than the luminance and the color of the radiation received by input image. Hence, an identification procedure can make it much more efficient which is based the detection of features that contains of some information about the classes which needed to be detected. Haar-like features is able to encode the presence of oriented contrasts among regions of the input image. One of the problems that this type of approaches presents is the computation time that is needed to calculate each of the features as a window sweeps the whole image at various scales.

## 3.4 Face Validity Process

The existing face detection methods have lack accuracy, especially in real time applications. Their results are not satisfactory because of their false alarm rate [3]. The problem is caused when face detectors wrongly select a non-face, which is known as false positives (FP). There is a need to decrease the number of false alarms due to most of the face detection algorithms having been used as the first stage in many applications. If this stage does not work well, the next stages cannot continue to work. There are some studies, which have been done to tackle that issue. Even though these studies could enhance the detection rate, the high number of false alarms are still a challenging issue, due to lack of any validation process to verify the selected face(s). Hence, a face validation process is proposed to retrieve face images again to recognize and filter out these false positives in order to improve the detection accuracy. The process aims to decrease the number of false alarms that may be created in previous stages. The input of this stage is the result of previous stage to verify the shape, face color skin and geometric characteristics of the input image. The performance is evaluated by the false alarm error (false positive rate) which is defined in Equation 2.

$$\gamma_{FA} = \frac{Number\ of\ false\ detection\ windows}{Total\ number\ of\ windows} \quad (2)$$

In this study, a complement stage is proposed with the aim of eliminating the false positives, which might be created in the proposed face detection. The algorithm is based on a linear Support Vector Machine (SVM) as a precise classifier in order to seek for a face in a two-dimensional solution space [7, 18]. In addition, for extracting features of potential faces, an extended Local Binary Pattern method is proposed. As a result, by employing the post-processing in the proposed framework, the number of false alarms will be decreased noticeably, which is the weakness of most existing face detection frameworks. The face validation process framework is based on the LBP, which consists of five main stages, including feature extraction, skin segmentation, Coarse Stage Feature Extraction, Fine Stage Feature Extraction, and classification, as depicted in Figure 5.

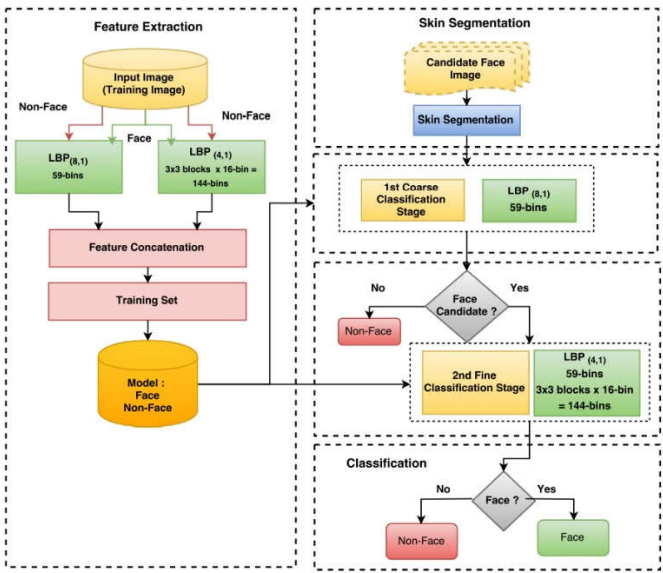

**Figure 5. The validity process design.**

The first stage of the framework confirms whether the global image appearance is potentially a face. Therefore, LBP (R = 1, N = 8) is extracted from the whole image obtaining a 59-bin labels histogram to have a precise explanation of the global image. Figure 6 depicts LBP (R = 1, N = 8). Only the potential faces, which passed the previous step, will be verified by means of a fine second stage that examines the spatial allocation of texture descriptors.

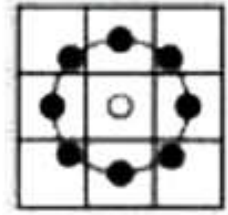

**Figure 6. LBP (R = 1, N = 8) Operator.**

For this reason, LBP (R = 1, N = 8) is applied to obtain entire 16×16 pixel image and 14×14 result images. After that, the result will be divided into 3×3 block of size 6×6 with 2 pixel overlapping. Figure 7 illustrates the process, which the first 6×6 pixel block indicated with gray color, and the rest of overlapping blocks are represented with red color lines.

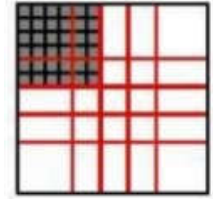

**Figure 7. Fine stage - 3×3 blocks division with two pixels overlapping.**

By using 3x3 blocks division for each block, will have a brief description of the local region by means of its 16-bin labels histogram. Consequently, an amount of 144 (3x3 x 16-bin) features vector is gained from the second fine stage. In addition, a different weight can be applied to each region in the classification stage in order to highlight most significant face regions. In the training stage, these two resulting histograms are concatenated with an enhanced feature vector, resulting in a face representation histogram of 59_bin + (3x3 blocks x 16_bin) = 59_bin + 144_bin = 203_bin. Figure 8 depicts the result of applying LBP in first Coarse and second fine stages.

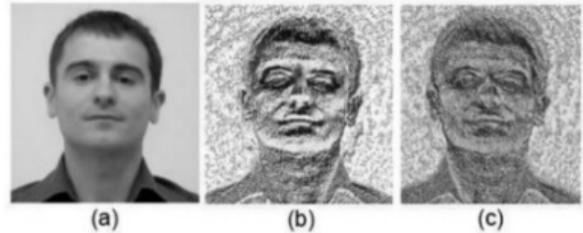

**Figure 8. Example of Image labeling (a) Original Image. (b) 1st stage $LBP_{(4,1)}$ (c) labels images rescaled to 0 …255.**

As illustrated in Figure 8, the contours of the facial features (mouth, eyes nose, ears, lips, etc.) are noticeably highlighted. In first stage labels image, contours are robustly highlighted and gives a superb overview of the image face object, which is helpful to distinguish between face and non-face images in first fast stage. Furthermore, in the second stage labels image, local texture information has more detailed which is helpful for final stage to decide between face and non-face objects.

## 4. RESULTS AND EVALUATIONS

In this section, the experimental results of the proposed framework are discussed. In addition, several well-known works in this area have been implemented to evaluate the performance of the proposed face detection framework. All of the experiments performed are on a standard dataset and compared with the implemented methods. The present section includes the proposed face detection framework results and related discussion along with the proposed skin color segmentation and validation methods.

### 4.1 Dataset

In order to implement and evaluate the experimental results with the existing methods, several standard datasets have been used, including the CMU + MIT face database, Caltech 100000 web face database, and GTAV image dataset. The standard face databases consist of a large number of images, which includes several variations.

### 4.2 Preprocessing Result

Noise removal is a significant phase of the image processing to improve the quality of a input image. In this stage, three filters and adjustments are performed to enhance the image quality before starting the detection algorithm. The preprocessing step consists of resolution reduction, median filtering, and histogram equalization. Resolution reduction is applied on the input image in order to reduce the dimensionality of the input images that can help to increase the computation time without losing too much information. The examination are performed on the Caltech 10000 dataset images within 5 different resolutions, starting from 640×480 pixels and then decreasing the image size 2, 4, 8 and 12 times. It can be inferred that the deduction of the image resolution even by 4 times does not affect the face detection process. The result shows that if the resolution deduction applied 8 times on the image, the detection rate would be decreased around 10% to 20%.

### 4.3 Search Space Reduction Result

In order to compare the proposed skin segmentation process with the existing methods, there is a distinct lack of infrastructure when it comes to the datasets, which are used by contributions made to the field of skin segmentation. Therefore, two existing methods have been implemented, and their performances are analyzed using a subset of images from identical images, which are used to evaluate the performance of proposed method. The proposed method has an average of around 69% reduction of search space and when compared to the selected methods. The proposed method has average of 3% enhancement in terms of search space reduction. In addition, in terms of evaluation, four different metrics in the analysis of segmentations are performed. The metrics of evaluations including Recall, Precision, Specificity, and Accuracy. Furthermore, all of them are based on the number of true positives (TP), true negatives (TN), false positives (FP), and false negatives (FN) yielded by the processing of a given image. The result of proposed method along with several selected approaches are provided (by setting a 0.5 threshold) in Table 1.

**Table 1. Comparison of proposed method on Caltech dataset.**

| Method | Recall | Precision | Specificity | Accuracy |
|--------|--------|-----------|-------------|----------|
| RGB | 74.19% | 49.88% | 81.92% | 79.65% |
| HSV | 66.30% | 43.53% | 80.06% | 78.07% |
| YCbCr | 73.70% | 53.37% | 85.44% | 83.05% |
| Proposed | 32.15% | 63.28% | 96.89% | 86.42% |

As can be seen from the result, the proposed segmentation method has better accuracy among the selected existing approaches. As a result, this process is very important in the proposed framework, due to non-skin objects being eliminated prior to the face detection process, which starts with the feature extraction from the selected regions from the previous stage.

### 4.4 Feature Extraction Result

In addition, we have compared the accuracy of using the basic Haar-like features and enhanced Haar features in different datasets, including the MIT dataset and CMU Frontal view dataset. The results show that the enhanced Haar-like features have better accuracy and performance than conventional Haar-like features on MIT and CMU-Profile datasets. The main reason is that the basic Haar features able to detect face in frontal view and images which contain profile and complex background cause high false alarms. Hence, in this study we have applied several rotated features to extract features of human face in both profile and frontal views in the standard face database images. This enhancement is because of the modified features of the Haar-like, which aims to extract the features of the human faces much accurately, particularly where the faces are not in frontal view [22].

### 4.5 Classification Result

The result of face detection that have been evaluated in order to receive the windows contain skin color regions which have been detected in the previous stage and the learning algorithm is applied to train the classifiers based on the given face and non-face objects. In addition, in order to obtain high detection rate (around 99%); we have set the maximum number of weak classifiers in different stages. In addition, the training stage may take 18 hours to be completed.

As the number of stages increases, the cascade focuses on hard samples, which have not been classified correctly by the previous stages. In this research, in the detection process, a cascade of 15 stages is applied for classification purposes. Over 55% of the examples are rejected in the first stage. The last stage of detection only classifies a few difficult-to-classify samples.. n=20 nodes have been used for classification. Surprisingly, 59.8% of samples have been rejected at the first stage. The last stage only struggled with few difficult-to-classify samples. The accuracy of the cascade Adaboost method and the validation has been evaluated and the number of Hits, Misses, False positives, and Detection rate are shown in Table 2.

**Table 2. Accuracy evaluation of Adaboost and ExLBP+SVM.**

| Detection Method | Hits | Misses | False positives | Detection rate (%) |
|------------------|------|--------|-----------------|--------------------|
| Adaboost Cascade | 5001 | 478 | 338 | 91 |
| ExLBP + SVM | 5046 | 451 | 97 | 91 |

Table 2 shows the accuracy evaluation of the cascade Adaboost method and the LBP+ SVM on a set of 5500 test images acquired from the Caltech 10000 Web Face database. The table compares the main face detection module and validation process (ExLBP + SVM) accuracy, separately, on 5500 sample test images from the Caltech 10000 Web Face database. As shown in this Table, the combination of the LBP and SVM method obtained higher detection rate with fewer false positives compared to cascade Adaboost method. Table 3 shows the evaluation result of the cascade Adaboost module, and the proposed method in terms of detection rate and respective number of false positives. The results show that in the same detection rate, the proposed method removed many of the false positives from the images that were received from the previous phase.

Table 3. Accuracy evaluation of Adaboost and the proposed method.

| Detection method | Hits | Misses | False positives | Detection rate (%) |
|---|---|---|---|---|
| Adaboost Cascade | 474 | 31 | 142 | 92 |
| Proposed method | 472 | 33 | 87 | 92 |

Table 3 shows the accuracy evaluation of the cascade Adaboost algorithm and the proposed method on the CMU+MIT test database containing 130 grey scale images with 507 face. As shown in the table, the number of false positives decreased significantly at the same detection rate compared with the proposed method without validation process. The result shows the effect of the verification unit to improve detection accuracy.

## 4.6 Framework Result

The operation of some face detectors and the proposed face detection system on CMU profile test set has been illustrated in Figure 9. As seen from the figure, the proposed method has high detection rate with a low number of false positives, which is significantly remarkable in comparison with selected existing methods. While the integral image ought to have direct usage for other systems, which use Haar-like features, it can foreseeably have effect on any task where Haar-like features may be of value. Preliminary experiments have presented that an akin feature set is also operative for the task of parameter approximation, where the appearance of a face, the location of a head, or the pose of an object is resolved.

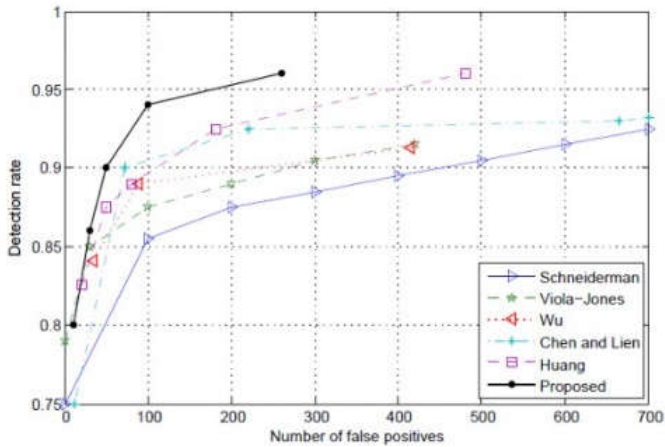

**Figure 9. ROC curve of face detection on CMU profile test set (208 images, 441 faces, 347of them are non-frontal).**

We implement two different methods in the validation process, compare each method, and report the system accuracy in types of faces data sets, and their ROC curves. Figures 10 and 11 illustrate the ROC curve for conventional LBP and ExLBP, respectively. The figures verify four parameters in order to get the statistic for the rate of true positive and false negative in a certain number of input images.

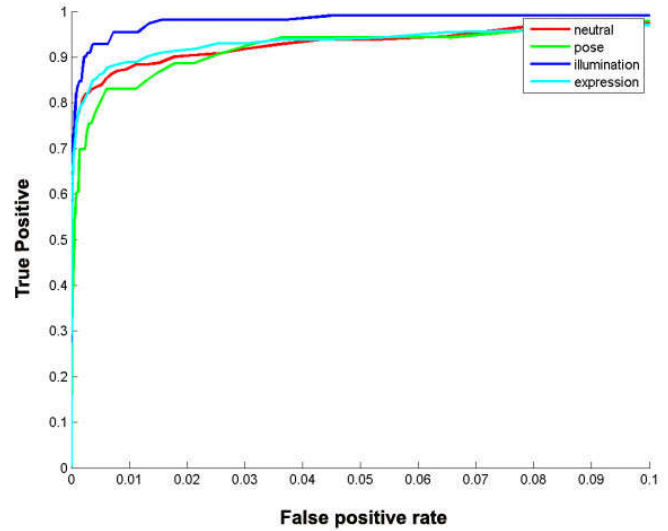

**Figure 10. ROC curve for LBP.**

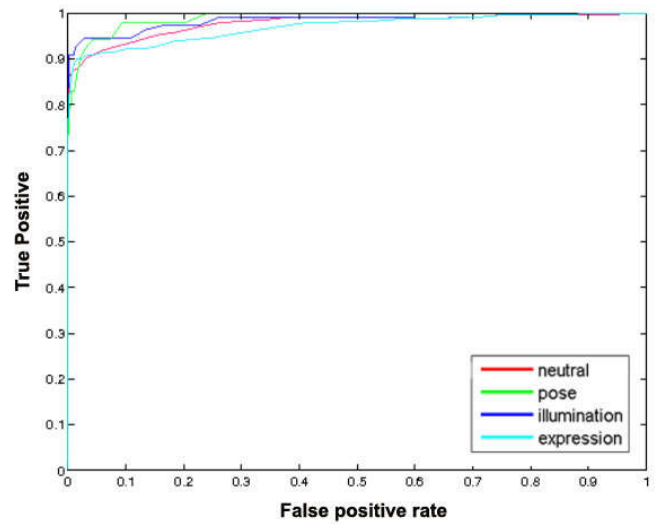

**Figure 11. ROC Curve for ExLBP.**

The proposed face detection framework has been implemented and the results of this framework are illustrated on several face databases including CMU-MIT, Caltech, and personal images as they are illustrated in Figure 12.

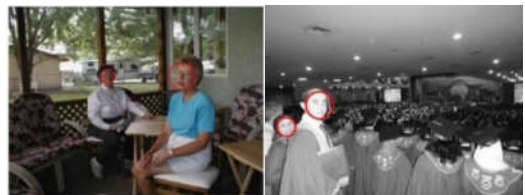

**Figure 12: Result of the proposed face detection (Frontal-view and Profile).**

## 5. CONCLUSION

The proposed face detection system consists of four main modules, including preprocessing, search space reduction, face detection, and the validation process. The preprocessing module aims to prepare the image before entering the face detection system in order to decrease the image size, enhance image quality, and remove noise. Then, the skin segmentation process is applied for search space reduction based on the skin color and edges of face(s) in the input images. Though, detecting skin-colored pixels seems to be a straightforward task, it has proven quite challenging for many reasons. The appearance of skin in an image depends on the lighting conditions where the image was captured. In the next stage, by using Haar-like algorithms, the features of input image are extracted based on the windows that consist of human faces. In addition, the Adaboost algorithm starts training the face and non-face images, which we have already given. The result of detection will contain human face images. To avoid this kind of false alarm, we have come up with the validation process. In the last module, the input is potential human face images, which needed to be validated. Finally, the selected face by the validation process until will be bounded throughout the input image. The combination of the four stated modules considered as the proposed framework are able to enhance an image in early stages and detect the face in frontal and profile views in detection of areas that consist of human skin color.